\documentclass[letterpaper]{article} 
\usepackage[preprint]{aaai2027}  
\usepackage[hyphens]{url}  
\usepackage{graphicx} 
\usepackage{natbib}  
\usepackage{caption} 
\usepackage{amsmath}

\newcommand{\extractllm}{\texttt{EXTRACT}}
\newcommand{\replyllm}{\texttt{REPLY}}

\usepackage{booktabs}

\title{FormBharo: Designing and Evaluating a Voice Agent for Conversational Form Filling in Rural India}

\author{
    Aman Dalmia,
    Sanskriti Midha,
    Jigar Doshi
}
\affiliations{
    Artpark, Indian Institute of Science, Bengaluru, Karnataka, India\\\{aman.dalmia, jigar, sanskriti\}@artpark.in
}

\begin{document}

\maketitle

\begin{abstract}
In India, almost every social benefit starts with a form, yet the people who need these benefits most are often unable to read or write. Reaching them requires a spoken conversation. Today that work falls to frontline health workers who enroll beneficiaries one at a time, a poor use of their stretched capacity. We built \textbf{FormBharo} (``fill the form'' in Hindi), a voice agent that fills a structured form over a phone call under tight latency and cost budgets by pairing Large Language Models (LLMs) with deterministic, rule-based validation and flow control. It is being piloted with ARMMAN, an NGO running large-scale maternal and child mobile-health programs in India, to enroll low-income, Hindi-speaking mothers in antenatal and postnatal care. To our knowledge, it is the first conversational voice agent piloted to fill an enrollment form for this population. We openly release \textbf{FormVoiceAgentBench}, a new benchmark pairing human-recorded Hindi audio with 3,760 multi-turn conversation tests across 960 simulated calls, to evaluate our agent's components (transcription, extraction, reply generation) and end-to-end form completion under real acoustic variations. Form completion drops by up to $\sim$41 percentage points when LLMs receive error-prone real-speech transcripts instead of reference transcripts. The rule-based controls recover many turn-level extraction errors, helping smaller, cheaper models match or surpass frontier models on form completion. Component-level performance does not predict end-to-end performance: GPT-5.5 leads turn-level extraction accuracy on reference transcripts (99.8\%) but ranks lower on form completion. Since errors both propagate and cancel across the pipeline, the optimal choice of models emerges only through end-to-end evaluation. Finally, no single model is the best across accuracy, cost, and latency at once, so we use a Pareto-based weighted-sum scalarization for selecting a deployable configuration that balances the three.
\end{abstract}

\begin{figure*}[t]
  \centering
  \includegraphics[width=\textwidth]{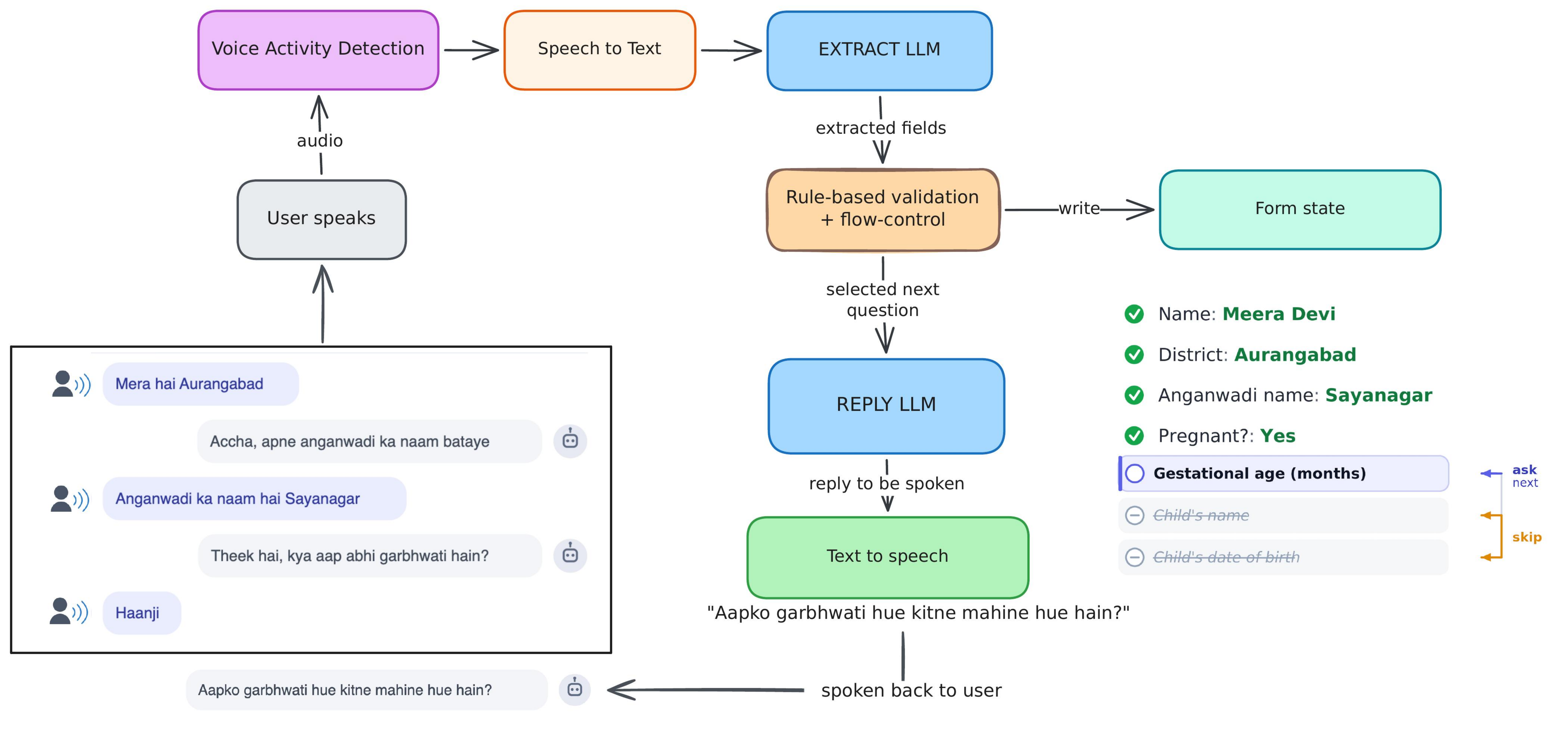}
  \caption{FormBharo's architecture. \extractllm{} extracts the form values from the transcribed text. The rule-based layer validates  the extracted values, updates the form state, skips inactive branches, and selects the next question to ask. \replyllm{} phrases the next question which is spoken back to the user through a TTS model. The ``Anganwadi name'' is the public clinic's name.}
  \label{fig:architecture}
\end{figure*}

\section{Introduction}
\label{sec:intro}

Access to nearly every social benefit in India runs through a form a citizen must complete to enroll \citep{IndusAction_2026}. Yet, the people with the greatest need for these programs are often disproportionately low-income and less literate, leaving them least equipped to access the benefits: more than half the women in the lowest wealth quintile cannot read at all (\citet{iips2021nfhs}).  Enrollment therefore falls to frontline health workers who sign up beneficiaries one conversation at a time \citep{dreze2017socialsecurity}. Documentation already consumes much of their day \citep{khandre2023timemotion} and dozens of overlapping reporting systems add to this burden, requiring the same data to be re-entered across apps and paper registers \citep{nongrum2025disconnected}. This caps enrollment at worker capacity rather than need. Additionally, these records are the administrator's source of truth for allocating resources, so errors and delays in enrollment slow the system's response.

Voice-based services have already demonstrated population-scale reach in low- and middle-income countries (LMICs). For example, the outbound prerecorded-call service Kilkari reached more than 10 million subscribers \citep{lefevre2022kilkari} and delivered $\sim$1.2 million calls per day \citep{bashingwa2021kilkari}. Early interactive voice response (IVR) systems commonly relied on touch-tone input, which users preferred over low-quality speech recognition \citep{patel2010avaaj}. However, \citet{sherwani2009speech} found that a carefully designed speech interface achieved significantly higher task-completion rates than an equivalent touch-tone interface among both low-literate and literate users. LLM-powered voice agents are now making real-time spoken interaction increasingly practical \citep{mukherjee-2026}, creating new opportunities for service delivery. A mother can call a number and enroll by speaking naturally with an agent in her own words and language, with nothing to read, type, or install. 

Doing this reliably is hard. Voice agents work well with a cooperative speaker in a quiet room, but performance degrades with background noise, telephone-channel distortion, underrepresented regional accents, unusual speaking rates, mispronunciations, and disfluencies \citep{chen2026voicebench,bhanushali2022gram}. Hindi-English code-mixing introduces another challenge \citep{diwan2021mucs}, as do structured values such as phone numbers and dates, which can be spoken in many ways but must resolve to a single value \citep{mohammadi-etal-2026-lingvarbench}. These problems compound in form-filling calls: callers hesitate, they correct themselves mid-answer, and the form branches on earlier answers, so an incorrectly captured field can send the agent down the wrong path.

We present \textbf{FormBharo}, a hybrid voice agent that fills a structured form over a phone call under tight latency and cost constraints (Figure~\ref{fig:architecture}). A speech-to-text (STT) model transcribes the caller's speech. An LLM, \extractllm{}, extracts the relevant form values. A rule-based layer then validates them, updates the form state and picks the next question to ask. A second LLM, \replyllm{}, phrases the question naturally, which a text-to-speech (TTS) model speaks back to the caller. If all the fields have been answered, the LLM decides to end the call instead. FormBharo is being piloted in rural Maharashtra, India, enrolling low-income, Hindi-speaking mothers in an antenatal and postnatal care program. The pilot runs in collaboration with ARMMAN~\cite{armman}, a nonprofit in India that operates large-scale mobile-health programs for maternal and child health among underserved communities.

To evaluate our agent, we release \textbf{FormVoiceAgentBench}, a benchmark pairing human-recorded Hindi audio with 3,760 multi-turn conversation tests across 960 simulated calls, built on the enrollment form from our pilot without exposing any real caller's data\footnote{Code and data will be made available here: \url{https://github.com/dalmia/AAAI-FormBharo-final/tree/main/code}}. Unit tests are used to evaluate the quality of transcription, data extraction, and reply generation individually. Integration tests chain them to measure end-to-end form completion with real-speech input.

Our experiments show that component-level performance does not predict form completion: LLMs that lead turn-level extraction accuracy with reference transcripts as inputs rank lower once the components are chained. The rule-based layer recovers many extraction errors, helping smaller, cheaper models match or surpass frontier models on form completion. Since errors both propagate and cancel across the pipeline, the best configuration emerges only through end-to-end evaluation. Finally, no single model is the best across accuracy, cost, and latency at once. To pick the best combination of models, we discard those that fail to satisfy our deployment constraints, keep the Pareto-optimal ones, and rank the rest by weighted-sum scalarization \citep{marler2010weighted}, with the weights reflecting our deployment's priorities.

Our contributions are summarized below:
\begin{itemize}
    \item We frame the call-based conversational form-filling task for low-literacy users in LMICs  and characterize the challenges that make it hard.
    \item We present \textbf{FormBharo}, to our knowledge the first conversational voice agent piloted to fill an enrollment form for this population.
    \item We share our evaluation design methodology and openly release \textbf{FormVoiceAgentBench}, a new benchmark in Hindi that implements it.
    \item We compare different model choices and share our findings from component-level and end-to-end evaluations across accuracy, latency, and cost.
\end{itemize}

\section{System Architecture}
\label{sec:architecture}

\textbf{FormBharo} is a voice agent that fills a structured form over a phone call by asking one question at a time. The design principle is to use LLMs only where they add value: interpreting user inputs, turning unstructured answers into structured fields, phrasing the next question naturally to be spoken back to the user, and deciding when the form is complete. Everything else stays rule-based: validation, retries, branching, and choosing the next question.

Figure~\ref{fig:architecture} shows the call flow. Once the Voice Activity Detection (VAD) model detects that the caller has stopped speaking, an STT model transcribes the audio. The \extractllm{} LLM reads the transcript together with the conversation history and extracts every field answered in that turn, so the agent does not re-ask questions already answered. The form has fields of several types: free text, number, date, boolean, and categorical. For categorical fields, \extractllm{} returns the index of the chosen option rather than the option text. In parallel, \extractllm{} generates a short acknowledgement that the TTS model speaks back to reduce perceived latency. 

The rule-based layer validates each extracted value against field-specific built-in guards, such as a specified minimum length, 10 digits for a phone number, or a date restricted to the past. When a value is missing or fails validation, it checks whether the maximum number of retries for that field is reached. If retries remain, the validation error is passed to \replyllm{} LLM along with the conversation history, which asks the question again. Once retries are exhausted for a required field, the rule-based layer ends the call. For an optional field, it either ends the call or skips it, depending on the field-specific setting. \extractllm{} also detects when the caller declines or does not know the answer to an optional field (the \emph{skip flag}). That field is then skipped without triggering a retry. When validation passes, the values are written to the form state, the single source of truth for which fields have been answered. The form has several branches conditioned on the values of earlier fields. For example, gestational age is asked only if the caller says she is pregnant. The rule-based flow-control evaluates whether any of the form's branches have been activated or disabled, and decides the next question to ask. This decision is injected into the conversation history of \replyllm{} LLM as a tool call. It phrases the question naturally for the TTS model to speak back to the caller, or ends the call if the form is complete.

Finally, if the user starts speaking while the agent's inference is running, the current turn is terminated immediately and the agent goes back to waiting for the user to stop speaking, so interruptions are handled gracefully.


\begin{table}[!htb]
\centering
\small
\setlength{\tabcolsep}{4pt}
\begin{tabular}{llc}
\hline
Field & Example answer & On fail \\
\hline
Name                      & Sunita Devi & skip \\
District                  & Mayurbhanj        & end  \\
Clinic name               & City Clinic & skip \\
Pregnant?                 & yes\,/\,no  & end  \\
\quad\textit{yes:}\ Gestational age      & 5 months    & end \\
\quad\textit{no:}\ Child's name          & Aarav       & end \\
\quad\phantom{\textit{no:}\ }Child's DOB & 15-04-2025  & end \\
Phone linked to clinic?   & yes\,/\,no  & skip \\
\quad\textit{no:}\ Number linked         & 9876543210  & skip \\
Phone linked to WhatsApp? & yes\,/\,no  & skip \\
\quad\textit{no:}\ WhatsApp number       & 9876543210  & skip \\
Aadhaar last 4 digits     & 4321     & skip \\
\hline
\end{tabular}
\caption{The enrollment form for testing FormBharo. When a caller fails to give a valid answer after a few retries, the agent either skips to the next field (\emph{skip}) or ends the call (\emph{end}). Indented fields are conditional: whether they are asked depends on the parent field's response. Aadhaar is India's national ID.}
\label{tab:form-fields}
\end{table}

\section{FormVoiceAgentBench}
\label{sec:benchmark}
\label{sec:dataset}

\begin{table}[t]
\centering
\small
\begin{tabular}{lr}
\hline
\textbf{Statistic} & \textbf{Count} \\
\hline
Simulated users & 5 \\
Form fields & 12 \\
Acoustic conditions per utterance & 4 \\
\hline
Unique audio recordings & 380 \\
Calls (unique call scripts) & 240 \\
Calls (with acoustic variations) & 960 \\
\hline
Unit tests (total) & 3{,}760 \\
\quad \extractllm{} & 1{,}880 \\
\quad \replyllm{} & 1{,}880 \\
\hline
\end{tabular}
\caption{\label{tab:dataset-stats} FormVoiceAgentBench statistics.}
\end{table}

\textbf{FormVoiceAgentBench} is a Hindi benchmark, grounded in the maternal and child health enrollment form from our pilot, that pairs 380 audio recordings with 3,760 multi-turn conversation tests across 960 simulated calls (Table~\ref{tab:dataset-stats}) without exposing any Personally Identifiable Information (PII) from real callers. It evaluates the agent at two levels. \textit{Unit tests} score each component in isolation: predicted transcripts against reference transcripts, extracted form data against the expected values, reply quality and ability to end the call given that the extraction was perfectly accurate. \textit{Integration tests} chain the components to measure end-to-end form completion.

\subsection{Form Structure}

The form has 12 fields (Table~\ref{tab:form-fields}) which follow a fixed order including conditional branching. If the caller is pregnant, the agent asks her gestational age. Otherwise, it asks her recently born child's name and date of birth. If her calling number is not linked to the clinic, the agent asks for the linked number; otherwise it skips ahead. Similarly, the agent asks for her WhatsApp number if it differs from her calling number. The other fields are always asked.

\subsection{Simulated Users and Transcripts}
We define five \emph{simulated user profiles}, each with fixed personal details, corresponding to the fields we intend to capture.

Each form field has three representations: the \emph{value}, the desired form entry and the ground truth for measuring form completion (``Sunita Devi''); the \emph{reference transcript}, a naturally spoken rendering of the value, phrased as a caller would say it in a real conversation (``My name is Sunita Devi''); and the \emph{recording} of that transcript. The value and the transcript are generated sequentially by an LLM, under field-specific constraints defined with ARMMAN: names and district names with phonetically hard spellings, and numbers with many spoken variations that make them hard to transcribe. District names were deliberately selected from outside the pilot state for adequate stress-testing. 

For optional fields, we generate two linguistic variants of the transcript: one where the caller responds with the value and the other where she says she does not know it. We use OpenAI's GPT-5.5 \citep{openai2026gpt55} for the generation.  

\subsection{Audio Data Collection}
Each transcript was recorded to mimic the conditions in a public clinic across three acoustic variations: background noise (ambient chatter and nearby speakers), microphone distance (close or far, chosen at random), and speaking pace (fast or slow, chosen at random). Combined with the ideal acoustic condition, this gives four recordings per transcript.

Five native Hindi speakers were selected to match the target demographic: all female, aged 18--35, drawn from two states (Uttar Pradesh and Maharashtra) to cover differences in accent and colloquialisms. Annotators were trained with sample recordings and recorded each answer under specific directives. For example, ``record in a noisy environment, or keep the mic at least 25\,cm away, or speak slowly''. A separate set of human supervisors listened to every clip to ensure the recordings met the requirements.

\subsection{From Transcripts to Calls}
We assemble each call by stitching together a simulated user's transcripts with one recording per field. Since the form branches, a simulated user can follow several paths, and each path becomes its own call. For example, the same simulated user produces one call in which she is pregnant and another in which she is not. Calls never mix values from different simulated users. Each call uses a single acoustic condition throughout (e.g.\ noisy environment or speaking slowly), since a caller's environment does not change mid-call.

Each simulated user follows $2\times3\times4\times2 = 48$ distinct paths through the form, one call per path: two pregnancy branches (pregnant or not), three branches for whether the calling number is linked to the clinic (linked, not linked and correct value provided, or not linked but not able to remember the linked number), four choices for whether the WhatsApp number is the same as the calling number (same, not same and answered, not same but unable to recall the WhatsApp number, cannot recall if they are the same), and two branches for the Aadhaar digits (answered or not known). Across the five simulated users this gives $5\times48 = 240$ calls, and recording each under the four acoustic conditions yields $240\times4 = 960$.

\section{Evaluation Design}
\label{sec:evaluation}

\subsection{Unit Tests}

\subsubsection{Speech-to-Text}

Word Error Rate (WER) is commonly used for comparing STT models. However, it is a poor metric for transcription quality on Indic languages \citep{sarvam2026asr} and for agents \citep{pipecat2026sttbench}: it counts every surface difference as an error, even when the meaning is unchanged. We therefore also report \texttt{LLM-WER} \citep{sarvam2026asr}, which discards mismatches that an LLM classifies as semantically equivalent or phonetically similar and recomputes WER over the genuine errors.

\subsubsection{\extractllm{} LLM}

To evaluate data extraction, we prepare unit tests from every call. For each turn, we create a unit test using the latest user response paired with the preceding conversation history as input and the expected form values as the ground truth. So, one call produces many tests. Deduplicating tests with identical inputs across all 960 calls yields 1,880 unit tests.

The extracted values for closed-ended fields like numbers, dates, and booleans are scored by computing the exact match with the expected values. The skip flag is scored the same way. Outputs for open-ended fields like names are harder to rate since the expected value can have many phonetic forms (``Lakshmi'' versus ``Laxmi''), so we use a calibrated binary LLM judge (details are in the Appendix) to evaluate them. Since \extractllm{} also generates a short open-ended acknowledgement, we use the LLM judge to evaluate its quality against a rubric. An example is provided in the Appendix. An accurate extraction passes both the exact-match checks and the LLM judgments. We report the mean \emph{extraction accuracy}.

\subsubsection{\replyllm{} LLM}
We similarly prepare unit tests from the calls to measure response accuracy, with two key differences. Along with the latest user response and the preceding conversation history, \replyllm{} additionally receives a tool call carrying the decision of the rule-based layer as an input too (Section \ref{sec:architecture}). For the unit tests, we construct this tool call by passing the expected extraction values through the rule-based validation and flow-control step, so that measuring response accuracy can be isolated from extraction errors.

Secondly, \replyllm{} either phrases the next question or ends the call when the form is complete. The end-call decision is a tool call, evaluated by exact match. Since the generated reply can be phrased in many equally correct ways, we again rely on calibrated binary LLM judges to rate them. However, unlike what we did for \extractllm{}, each reply is graded across five independent dimensions: 1) asking the right question, 2) adherence to Hindi, 3) single line response, 4) no acknowledgment (since \extractllm{} handles that), and 5) not echoing the caller's answer back. The LLM judge calibration details and an example are in the Appendix. An accurate response either passes all the LLM judgments or ends the call at the right time. We report the mean \emph{response accuracy}.

As with \extractllm{}, deduplicating the tests across the 960 simulated calls, each call producing many tests, yields 1,880 unit tests: 920 require a reply, while the remaining 960 check whether \replyllm{} ends the call correctly.

\subsection{Integration Tests}

Integration tests chain the components, so the output of one feeds into the next, and the agent is judged on completing the form, not on any single turn.

First, we replace the reference transcripts in the unit-test inputs with transcripts produced by the STT models, so transcription errors propagate to the LLMs. Next, we evaluate each call as a whole. We start from an empty form and go through the call's turns in order. At each turn, we write the values extracted for that turn's unit test into the form, as the agent would in a live call. The form after the last turn is the predicted final form. Comparing this against the expected values gives the \textit{form-completion accuracy}: the fraction of form fields captured correctly at the end of the call. We compute it using both reference transcripts and the transcripts produced by each STT model as inputs. The gap between them quantifies the impact of transcription errors on form completion. Finally, we chain \extractllm{} and \replyllm{}. To build the tool call for \replyllm{}, we pass \extractllm{}'s actual output through the rule-based validation and flow control, instead of the expected extraction values used in the unit tests.

Each \extractllm{} model drives its own set of \replyllm{} runs. Because each stage feeds the next, extraction accuracy is measured for every STT $\times$ \extractllm{} pair, and response accuracy for every STT $\times$ \extractllm{} $\times$ \replyllm{} combination.

\section{Experiments \& Analysis}
\label{sec:experiments}

\begin{table}[t]
\centering
\small
\setlength{\tabcolsep}{6pt}
\begin{tabular}{lrrr}
\hline
Model & WER $\downarrow$ & LLM-WER $\downarrow$ & Cost (USD) $\downarrow$ \\
\hline
Scribe v2 & $0.835$ & $0.062$ & $0.172$ \\
Chirp 3 & $0.981$ & $\mathbf{0.055}$ & $0.424$ \\
Saaras v3 & $1.007$ & $0.074$ & $0.137$ \\
GPT-4o-transcribe & $1.012$ & $0.127$ & $0.159$ \\
Nova-3 & $\mathbf{0.826}$ & $0.104$ & $\mathbf{0.127}$ \\
\hline
\end{tabular}
\caption{Comparison of STT models. Since models have different billing units, the total cost of transcribing the entire benchmark is reported.}
\label{tab:stt-cost}
\end{table}

\subsection{Setup}

We benchmark five STT models and 11 LLMs across accuracy, latency (p95), and cost. The temperature is set to 0 for non-reasoning models. The reasoning models use ``medium'' reasoning effort for \extractllm{} and ``low'' for \replyllm{}. Full implementation details and 95\% confidence intervals can be found in the appendix.

\subsection{Speech-to-Text}

Table~\ref{tab:stt-cost} shows that the two metrics disagree: Nova-3 has the best WER yet the second-worst \texttt{LLM-WER}. We therefore rank models by \texttt{LLM-WER}, which was built to address the shortcomings of WER. Chirp 3 has the best \texttt{LLM-WER} but costs at least twice as much as any other model, while Scribe v2 performs close to Chirp 3 at a fraction of the cost. GPT-4o-transcribe is the least accurate. Eliminating these two leaves Scribe v2, Saaras v3, and Nova-3, which we carry into the integration tests.

\begin{table}[t]
\centering
\small
\setlength{\tabcolsep}{3pt}
\begin{tabular}{l cccc}
\hline
Model & Reference & Saaras v3 & Scribe v2 & Nova-3\\
\hline
GPT-5.5 & $\mathbf{99.79}$ & $95.53$ & $98.62$ & $95.16$\\
Gemini 3.5 Flash & $99.36$ & $95.69$ & $\mathbf{98.94}$ & $\mathbf{96.38}$\\
Claude Opus 4.8 & $99.15$ & $\mathbf{96.22}$ & $98.40$ & $96.01$\\
Claude Sonnet 4.6 & $99.15$ & $96.01$ & $\mathbf{98.94}$ & $95.96$\\
GLM-5.1 & $98.72$ & $94.52$ & $63.40$ & $95.43$\\
Gemini Pro & $97.23$ & $94.26$ & $97.29$ & $94.63$\\
Gemini 2.5 Flash & $96.17$ & $95.27$ & $98.03$ & $95.43$\\
Gemini 3 Flash & $95.96$ & $84.79$ & $88.83$ & $86.28$\\
GPT-5.4-mini & $92.34$ & $90.05$ & $96.76$ & $91.54$\\
Mistral Medium 3.5 & $89.36$ & $87.82$ & $92.34$ & $87.77$\\
GPT-4.1 & $85.11$ & $75.53$ & $78.94$ & $79.84$\\
\hline
\end{tabular}
\caption{Per-turn extraction accuracy using reference transcripts and the transcripts from the three STT models as inputs.}\label{tab:extract-only}
\end{table}

\begin{table}[t]
\centering
\small
\setlength{\tabcolsep}{3pt}
\begin{tabular}{l cccc}
\hline
Model & Reference & Saaras v3 & Scribe v2 & Nova-3\\
\hline
Gemini 3 Flash & $\mathbf{100.00}$ & $85.24$ & $89.48$ & $83.92$\\
Gemini 3.5 Flash & $\mathbf{100.00}$ & $90.74$ & $92.50$ & $\mathbf{87.09}$\\
GPT-5.5 & $99.95$ & $89.97$ & $89.37$ & $84.73$\\
Claude Sonnet 4.6 & $99.66$ & $91.55$ & $93.01$ & $84.54$\\
GLM-5.1 & $99.56$ & $90.92$ & $58.66$ & $86.09$\\
Claude Opus 4.8 & $99.41$ & $92.22$ & $92.03$ & $85.83$\\
Gemini 2.5 Flash & $99.06$ & $91.76$ & $92.39$ & $85.50$\\
Gemini Pro & $98.90$ & $\mathbf{92.55}$ & $\mathbf{93.47}$ & $86.53$\\
GPT-5.4-mini & $98.35$ & $90.68$ & $90.10$ & $84.40$\\
GPT-4.1 & $95.99$ & $88.82$ & $91.50$ & $86.33$\\
Mistral Medium 3.5 & $92.83$ & $90.39$ & $91.84$ & $83.37$\\
\hline
\end{tabular}
\caption{Form-completion accuracy using reference transcripts and the outputs of the three STT models as inputs.}
\label{tab:form-completion-only}
\end{table}

\begin{table}[t]
\centering
\small
\setlength{\tabcolsep}{3pt}
\begin{tabular}{lcccc}
\toprule
& \multicolumn{2}{c}{Extraction accuracy}
& \multicolumn{2}{c}{Form completion} \\
\cmidrule(lr){2-3}\cmidrule(lr){4-5}
Model & Median (\%) & $\Delta$ (pp) & Median (\%) & $\Delta$ (pp) \\
\midrule
Saaras v3 & $94.52$ & $-3.14$ & $90.74$ & $-7.67$ \\
Scribe v2 & $\mathbf{97.29}$ & $\mathbf{-0.42}$ & $\mathbf{91.84}$ & $\mathbf{-7.38}$ \\
Nova-3 & $95.16$ & $-3.14$ & $85.50$ & $-13.56$ \\
\bottomrule
\end{tabular}
\caption{Median extraction and form-completion accuracy with real-speech transcripts as inputs. $\Delta$ denotes the median change relative to reference transcripts. Both degrade under transcription noise.}
\label{tab:stt-drop}
\end{table}

\subsection{Data Extraction}

Table~\ref{tab:extract-only} compares various LLMs on per-turn extraction accuracy, computed using reference transcripts and the transcripts of the three STT models as inputs. 

Frontier models saturate extraction accuracy on reference transcripts as inputs with the top models scoring almost perfectly: GPT-5.5 leads at 99.79\%, with Gemini 3.5 Flash (99.36\%) and the two Claude models (99.15\%) just behind. Robustness to STT errors varies across LLMs: the median drop is modest, from 0.42 percentage points to 3.14 (Table~\ref{tab:stt-drop}). The four most accurate models lose at most 4.63 points regardless of the STT model. But weaker models degrade much more: GLM-5.1 collapses by 35 points with Scribe v2 transcripts. The next-worst drop is still 11.2  points. The extraction accuracy leaderboard changes when real transcripts are used: GPT-5.5 is no longer the winner. The best extraction accuracy under real-speech is 98.94\% (Gemini 3.5 Flash and Claude Sonnet 4.6) with transcripts from Scribe v2. Claude Opus 4.8 leads under Saaras v3 (96.22\%) and Gemini 3.5 Flash leads under Nova-3 transcripts (96.38\%).

\subsection{Form Completion}
Similar to extraction accuracy, Table~\ref{tab:form-completion-only} reports form-completion accuracy across all the LLMs being tested. The rule-based layer recovers extraction errors. 

On reference transcripts, Gemini 3 Flash achieves 95.96\% per-turn extraction accuracy but 100\% form-completion accuracy. Per-turn extraction evaluates the output of \extractllm{} LLM, whereas form completion evaluates the values ultimately stored in the form after the rule-based layer processes it. All the extraction errors for this model arise from a type mismatch: a numeric field being returned as a string. The rule-based layer normalizes it before storing the value, preserving 100\% form-completion accuracy. This highlights a benefit of our hybrid design which enables smaller models to perform better end-to-end even if their per-turn inference is not perfect. 

The best model end-to-end differs from the best model per-turn. Under reference transcripts, GPT-5.5 leads extraction accuracy (99.79\%), whereas Gemini 3 Flash and Gemini 3.5 Flash tie for the highest form-completion accuracy at 100\%. 

Form completion degrades more than per-turn extraction with error-prone real-speech transcripts: median form-completion accuracy drops by 7.67 percentage points with Saaras v3, 7.38 points with Scribe v2, and 13.56 points with Nova-3 (Table~\ref{tab:stt-drop}), compared with median per-turn extraction drops of only 0.42--3.14 points. The largest model-specific decline is $\sim$41 points for GLM-5.1 with Scribe v2. Although the rule-based layer recovers some extraction errors, uncorrected errors can accumulate across turns to produce a larger degradation end-to-end.

\subsection{Model Selection}

We select the STT, \extractllm{}, and \replyllm{} models in the pipeline (Figure~\ref{fig:architecture}) sequentially because each downstream component consumes the outputs of the components before it. We first select the STT model, then identify the best LLM for \extractllm{} using that STT model's transcripts, and finally evaluate \replyllm{} with the selected STT and \extractllm{} models fixed. The deployment objective is to balance task performance, p95 latency, and cost, subject to component-specific deployment constraints.

\textbf{STT selection.} Scribe v2 provides the strongest downstream performance. It achieves the highest median extraction accuracy (97.29\%) and form-completion accuracy (91.84\%), while producing the smallest median drops relative to the model performance on reference transcripts (Table~\ref{tab:stt-drop}). We therefore select Scribe v2 as the STT model.

\begin{figure}[!ht]
\centering
\includegraphics[width=\columnwidth]{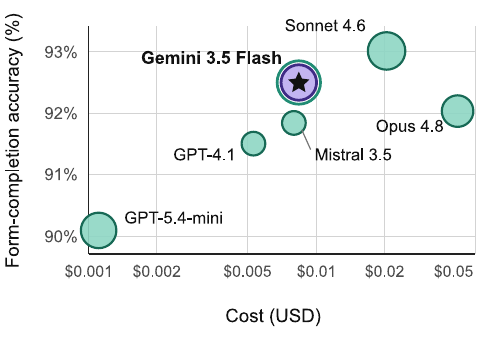}
\caption{Cost--quality--latency trade-off among \extractllm{} models satisfying the deployment constraints. The logarithmic x-axis shows cost per turn, the y-axis shows form-completion accuracy using Scribe v2 transcripts, and marker area encodes p95 latency per turn (bigger is slower). All six models are Pareto-optimal across the three objectives. The model selected for deployment is highlighted.}
\label{fig:model-selection}
\end{figure}

\textbf{\extractllm{} selection.} With Scribe v2 fixed as the STT model, we compare \extractllm{} models using form-completion accuracy on its transcripts, together with latency and cost. We discard models that fail our deployment constraints: p95 latency below $5$\,s and form-completion accuracy above $90\%$. This excludes Gemini Pro, despite its leading form-completion accuracy, and Gemini 2.5 Flash, leaving behind six candidates. Among those, Claude Sonnet 4.6 achieves the highest form-completion accuracy, Mistral Medium 3.5 has the lowest p95 latency (1.75\,s), and GPT-5.4-mini is the cheapest (\$0.0011). No model leads all three axes, and all six candidates lie on the Pareto frontier (Figure~\ref{fig:model-selection}), so selecting a deployable model requires balancing these competing objectives. First, we min--max normalize the three axes within the frontier. Since lower latency and cost are better, we reverse their scales so that higher values are always preferred:

\[
\begin{gathered}
\tilde{a}_i=\frac{a_i-a_{\min}}{a_{\max}-a_{\min}},
\qquad
\tilde{\ell}_i=\frac{\ell_{\max}-\ell_i}{\ell_{\max}-\ell_{\min}},\\
\tilde{c}_i=\frac{c_{\max}-c_i}{c_{\max}-c_{\min}}.
\end{gathered}
\]

Next, we rank the models using weighted-sum scalarization~\citep{marler2010weighted}:
\[
U_i = w_a\tilde a_i + w_\ell\tilde\ell_i + w_c\tilde c_i,
\qquad
i^\star=\arg\max_i U_i,
\]
where $w_a+w_\ell+w_c=1$. These weights can be tuned to reflect the deployment priorities.

As a baseline, under equal weights, Mistral Medium 3.5 ranks highest ($U=0.81$), followed by GPT-4.1 ($U=0.78$). Mistral combines the lowest latency with 91.84\% form-completion accuracy and a cost of \$0.0080 per turn. Claude Sonnet 4.6, the most accurate candidate, scores lower ($U=0.51$) because it is slower and more expensive. The hard latency constraint already excludes models that are too slow for deployment, so we assign the remaining latency differences a lower weight, $w_\ell=0.1$. Because accuracy remains the primary objective, we sweep $w_a$ from $0.5$ to $0.9$, with $w_c=1-w_a-w_\ell$. Only two models lead across this range: Gemini 3.5 Flash for $0.50 \leq w_a \leq 0.61$, and Claude Sonnet 4.6 for $0.62 \leq w_a \leq 0.90$ (full table in the Appendix). We select Gemini 3.5 Flash for \extractllm{}: it trails Claude Sonnet 4.6 by only 0.51 percentage points on form-completion accuracy while responding faster and costing less than half as much per turn.

\subsection{Response Generation}

\begin{table}[t]
\centering
\small
\setlength{\tabcolsep}{8pt}
\begin{tabular}{@{}lcc@{}}
\toprule
Model & Reference & Scribe v2 \\
\midrule
Claude Sonnet 4.6 & $\mathbf{100.00}$ & $\mathbf{97.77}$ \\
GPT-4.1 & $\mathbf{100.00}$ & $96.60$ \\
GPT-5.4-mini & $98.09$ & $96.70$ \\
Gemini 3 Flash & $97.23$ & $96.17$ \\
Gemini 3.5 Flash & $97.02$ & $94.73$ \\
\bottomrule
\end{tabular}
\caption{\replyllm{} response accuracy with Gemini 3.5 Flash as the \extractllm{} LLM.}
\label{tab:reply}
\end{table}

\replyllm{} has the narrowest role in the pipeline (Section~\ref{sec:architecture}): it either phrases the selected question naturally or ends the call. So, a smaller, faster model may suffice. With Scribe v2 and Gemini 3.5 Flash for transcription and extraction, we compare five LLMs for \replyllm{}. The selected \extractllm{} model's actual outputs on Scribe v2 transcripts are passed through the rule-based layer to construct the decision passed to \replyllm{}. Table~\ref{tab:reply} reports the resulting response accuracy. The Reference column is computed using the expected extraction values with reference transcripts to prepare the inputs, whereas for Scribe v2, the outputs of \extractllm{} on Scribe v2 transcripts are used to prepare the tests (Section~\ref{sec:evaluation}). 

Errors propagate through the pipeline. Response accuracy decreases for all five models when errors from transcription and extraction flow through to \replyllm{}. The decline ranges from 1.06 to 3.40 percentage points, with a median of 2.23 points, capturing the combined effect of transcription and extraction errors. No single model is the best across accuracy, latency, and cost. Claude Sonnet 4.6 is the most accurate on Scribe v2 transcripts (97.77\%), Gemini 3 Flash has the lowest p95 latency (2.66\,s), and GPT-5.4-mini is the cheapest (\$0.0008 per turn). Our deployment constraints of p95 latency below $4$\,s and response accuracy above $95\%$ exclude Claude Sonnet 4.6 on latency and Gemini 3.5 Flash on accuracy, leaving three candidates. Pareto filtering leaves GPT-5.4-mini and Gemini 3 Flash. Using the same weighted-sum scalarization method defined earlier with $w_\ell=0.1$, GPT-5.4-mini ranks highest throughout $0.5 \leq w_a \leq 0.9$ (details are in the Appendix). Therefore, we select GPT-5.4-mini for \replyllm{}.

\section{Related Work}
\label{sec:related}




 \textbf{Deployed AI for public services.} A growing body of AI-for-social-impact research studies how AI reshapes access to services for under-served users. \citet{jo2025aitrust} show LLM assistants can lower administrative burdens while adding new compliance and trust costs, and studies of digital welfare systems document similar transfers of burden to claimants \citep{watson2024precarious}.  Closest to our design, \citet{kothari2026safetynet} decompose a clinical LLM-summarization task into semi-structured attributes that can be validated separately rather than trusting one end-to-end prompt. This matters for equity: \citet{pooledayan2026vulnerable} found that the LLM's response quality drops for users with lower English proficiency and literacy. Our callers fit that profile, so any non-English system needs to be evaluated rigorously. 
 

\textbf{Spoken understanding and Indic speech.} Spoken slot filling and dialogue state tracking often use an STT-to-LLM cascade that transcribes and extracts values, where recognition errors propagate into slot and state errors \citep{yoon2023adapting, jacqmin2023olisia, ganesan2021n, sun2024speech}; \citet{si2023spokenwoz} show that a low WER does not guarantee task accuracy. Indic and code-mixed resources supply realistic acoustic and linguistic variation, including multilingual, code-switched, spontaneous, telephonic, geographically diverse, and low-resource speech \citep{diwan2021mucs, bhanushali2022gram, bhogale2026voice, javed2022indicsuperb, javed2024indicvoices, vaani2026, joshi2025sruti}. However, these resources evaluate transcriptions or other component-level speech tasks such as speaker identification. In our paper, in addition to transcriptions, we score per-field correctness on a real Indian enrollment form.

\textbf{Voice agents  and form-filling.} Recent benchmarks such as VoiceBench \citep{chen2026voicebench} and VoiceAgentBench \citep{jain2025voiceagentbench} evaluate voice systems on outcomes beyond transcription such as spoken question answering, instruction following and tool selection, using predominantly synthetic speech. Closest to us, EVA-Bench evaluates task accuracy and interaction quality over simulated multi-turn enterprise calls \citep{bogavelli2026evabench}. None of them, however, target a constrained, structured task like form completion. 

Related application systems assess latency and conversational quality, form usability, clinician-reviewed speech to EMR generation, or sampled production records \citep{cuadra2024digital,mustafa2025systemx,mukherjee2026perfecting}. In contrast, \textit{FormVoiceAgentBench} scores many models on form completion using noisy audio in a low-resource language.


\section{Conclusion}
\label{sec:conclusion}

We presented FormBharo, a phone-call-based conversational form-filling voice agent being piloted in a live maternal and child health enrollment program in rural Maharashtra, India, and described FormVoiceAgentBench, a benchmark for evaluating it, along with our findings. Our results show that component-level accuracy does not predict end-to-end form completion, with errors both propagating and canceling across the pipeline. The rule-based layer recovers many model errors, helping smaller, cheaper models meet deployment constraints. This is critical in LMICs, where cost and latency constrain deployment at scale.

The current benchmark is limited in several ways. It captures scripted, well-formed answers, but real callers also give wrong, partial, or self-corrected values. Each call applies only a single acoustic variation at a time, so combinations within the same call, such as a distant microphone in a noisy room, remain untested. The benchmark is based on five simulated users, with audio recorded one turn at a time by five annotators from two states rather than through full live calls, limiting its conversational, linguistic, and demographic diversity. The dataset is also limited to Hindi, though our intended users are multilingual. Finally, we did not evaluate TTS output quality.


\section*{Ethical Statement}
\label{sec:ethics}

No real user data was used to build the dataset. The spoken scripts were recorded by paid annotators who are native Hindi speakers. The broader intended impact of FormBharo is to widen access to care for an under-served population, but this also increases the risks. Since incorrect data capture could instead delay or deny access, deployment at scale requires more rigorous testing with adequate guardrails and fallback mechanisms in place to confirm or correct captured information when required.

\section*{Acknowledgments}
We thank Amrita Mahale, Parina Anand and Hetvi Lodaya at ARMMAN for designing the enrollment form, testing the agent through successive iterations, and sharing the insights from the field that guided its design. We thank the Vaani team at ARTPARK for their help with data collection and annotation. Finally, we are grateful to the frontline health workers and  mothers who tested FormBharo and shared their feedback.

\bibliography{aaai2027}


\appendix


\section{Enrollment Form}
\label{app:form}
\begin{figure*}[p]
  \centering
  \includegraphics[width=\textwidth]{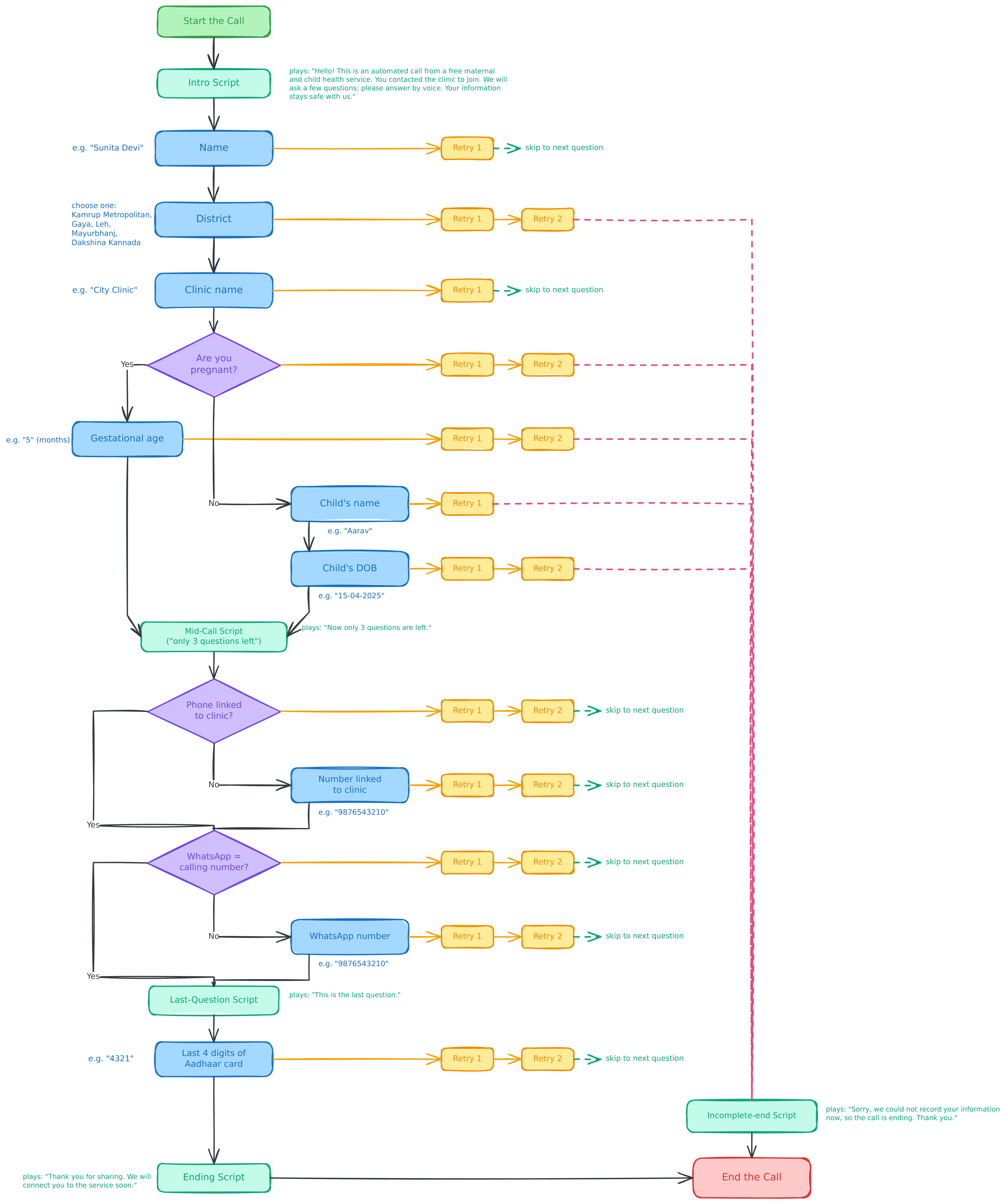}
\caption{Call flow of the enrollment form used to build \textbf{FormVoiceAgentBench}. The agent speaks first and asks one question at a time, in the order shown. The answers to the branching questions decide which of the conditional fields are asked next. If a caller gives no valid answer, the rule-based layer re-prompts up to the retry limit shown for that field, and then either skips to the next question or ends the call marking it incomplete. Aadhaar is India's national ID, of which only the last four digits are collected.}
  \label{fig:form-flow}
\end{figure*}

The full call flow of the enrollment form is shown in Figure~\ref{fig:form-flow}.

\section{LLM Judges}
\label{app:judges}

All LLM judges receive the conversation history, together with the agent output being graded, as the input, and produce a boolean score indicating whether the agent's output adheres to the judge criteria, following the LLM-as-a-judge paradigm \citep{zheng2023judging}. The temperature is set to $0$ for every LLM judge inference.

\subsection{\extractllm{} Judges}
\label{app:judge-extract}

Each extracted field is scored either by exact match or by an LLM judge (Table~\ref{app:field-match-type}), as is the acknowledgement the agent emits alongside the extraction. \texttt{openai/gpt-5.4-mini} is used as the judge model. The following instructions are added as the system prompt for the LLM judge, in which \texttt{\{\{criteria\}\}} is replaced by the criteria for the field being scored:

\begin{quote}
You are a highly accurate evaluator checking whether the value an agent produced for a single tool-call argument satisfies a given criteria.

You will be given the tool name, the argument name, and the actual value the agent produced for that argument.

Mark \texttt{match} true only if the actual value satisfies the following criteria, and false otherwise:

\texttt{\{\{criteria\}\}}
\end{quote}

An example is given below.

\begin{table}[t]
\centering
\small
\begin{tabular}{ll}
\hline
Field & Scoring \\
\hline
Name & LLM judge \\
District & exact \\
Clinic name & LLM judge \\
Child's name & LLM judge \\
Pregnant? & exact \\
Gestational age & exact \\
Child's DOB & exact \\
Calling number linked to clinic? & exact \\
Number linked to clinic & exact \\
WhatsApp = calling number? & exact \\
WhatsApp number & exact \\
Aadhaar last 4 digits & exact \\
\hline
\end{tabular}
\caption{How each form field is scored during \extractllm{} evaluation. Open-ended text fields are evaluated using an LLM judge, whereas every closed-ended field is scored by exact match.}
\label{app:field-match-type}
\end{table}

\begin{quote}
\textbf{Conversation history}

\textbf{assistant:} Namaste! Yah ek automated call hai, ek nishulk maatru evam shishu swasthya seva ki taraf se. Aapne hamari seva se judne ke liye Anganwadi se sampark kiya hai. Yah ek muft seva hai, jismein aapko maa aur bachche ke swasthya se judi upyogi jaankari milegi. Hamari seva se judne ke liye hum aapse kuch sawal poochhenge. Kripya unke jawab bolkar dein. Aapki di gayi saari jaankari hamare saath surakshit rahegi. Apna poora naam bataiye \emph{(Namaste! This is an automated call from a free maternal and child health service. You contacted the Anganwadi to join our service. This is a free service through which you will receive useful information about mother and child health. To enroll you in our service we will ask you a few questions. Please answer them by speaking. All the information you give will stay safe with us. Please tell me your full name.)}

\textbf{user:} Mera poora naam Mrinmayee Kshirsagar hai \emph{(My full name is Mrinmayee Kshirsagar.)}

\medskip
\textbf{Evaluation.} The turn is scored on the extraction call the agent makes. One field, the caller's name, is open-ended and so is sent to an LLM judge with the criteria below; the acknowledgement is judged against a rubric, since it has no single correct answer. Both criteria follow.

\medskip
\emph{Acknowledgement criteria.} A brief acknowledgement of the user's previous answer, made of real words (not a non-lexical sound, grunt, or filler). This is encompassing and NOT restricted to any specific words: accept a receipt, back-channel, reassurance, praise, or transition phrase in any language, for example (but not strictly limited to) \emph{thik hai}, \emph{achcha}, \emph{ji haan}, \emph{samajh gaya}, \emph{koi baat nahi}, ``okay'', ``yes'', ``alright'', ``got it'', ``thank you'', ``no problem''. Do not require any particular phrase, a `hearing-only' tone, or extra brevity; phrases that imply the agent understood or accepted the answer (e.g.\ \emph{samajh gaya}, ``got it'') are acceptable. Only a missing acknowledgement, or a non-lexical sound/filler, should fail this criterion.

\medskip
\emph{Name criteria.} The captured value refers to the SAME name as ``Mrinmayee Kshirsagar'' when read aloud. Treat them as EQUAL when they differ only by: letter casing; leading/trailing or internal word spacing; or a reasonable alternative romanisation / transliteration of the same spoken Hindi name, for example `ksh' vs `ksha', doubled vs single consonants, an inserted or dropped short `a'/schwa vowel, `v' vs `w', or `s' vs `sh' (so `Ghorakshpalli', `Ghorakshapalli' and `Goraksha Palli' are all equal). Mark it NOT equal only if it is a genuinely different name or refers to a different place/word, not for spelling or casing variants of the same name.
\end{quote}

The field scored above, the caller's name, is open-ended: the same spoken answer can be written many ways (for example, ``Lakshmi'' or ``Laxmi''). An LLM judge therefore decides whether the extracted value matches a given rubric for that field. ``Anganwadi'' refers to the local clinic.

A closed-ended field has one correct value and the extracted value is compared directly to it. The example below illustrates this.

\begin{quote}
\textbf{Conversation history}

\textbf{assistant:} Namaste! Yah ek automated call hai, ek nishulk maatru evam shishu swasthya seva ki taraf se. Aapne hamari seva se judne ke liye Anganwadi se sampark kiya hai. Yah ek muft seva hai, jismein aapko maa aur bachche ke swasthya se judi upyogi jaankari milegi. Hamari seva se judne ke liye hum aapse kuch sawal poochhenge. Kripya unke jawab bolkar dein. Aapki di gayi saari jaankari hamare saath surakshit rahegi. Apna poora naam bataiye \emph{(Namaste! This is an automated call from a free maternal
and child health service. You contacted the Anganwadi to join our service. This is a free service through which you will receive useful information about mother and child health. To enroll you in our service we will ask you a few questions. Please answer them by speaking. All the information you give will stay safe with us. Please tell me your full name.)}

\textbf{user:} Mera poora naam Mrinmayee Kshirsagar hai \emph{(My full name is Mrinmayee Kshirsagar.)}

\textbf{assistant:} Theek hai. Apne zile ka naam bataayein \emph{(All right. Please tell me the name of your district.)}

\textbf{user:} Form ke liye, mera zila Kamrup Metropolitan hai. \emph{(For the form, my district is Kamrup Metropolitan.)}

\textbf{assistant:} Theek hai. Apne Anganwadi ka naam bataayein \emph{(All right. Please tell me the name of your Anganwadi.)}

\textbf{user:} Mere aanganwadi ka naam Ghorakshpalli hai \emph{(The name of my Anganwadi is Ghorakshpalli.)}

\textbf{assistant:} Theek hai. Kya aap abhi garbhvati hain? \emph{(All right. Are you currently pregnant?)}

\textbf{user:} Haan main abhi garbhvati hoon. \emph{(Yes, I am currently pregnant.)}

\textbf{assistant:} Theek hai. Aapko garbhvati hue kitne mahine hue hain? \emph{(All right. How many months pregnant are you?)}

\textbf{user:} Abhi meri pregnancy teen mahine ki hai. \emph{(My pregnancy is three months along right now.)}

\medskip
\textbf{Evaluation.} The extracted value for how long the user has been pregnant must be $3$ (months). The acknowledgement is judged against the same rubric as before.
\end{quote}

\subsection{\replyllm{} Judges}
\label{app:judge-reply}

Each reply is graded independently on five dimensions, each using a separate LLM judge:

\begin{quote}
\textbf{Correctness} (\texttt{openai/gpt-5.4-mini}): does the reply ask the question the rule-based layer selected?

\textbf{Hindi Language adherence} (\texttt{openai/gpt-5.5}): is the reply in Hindi?

\textbf{Reply Conciseness} (\texttt{openai/gpt-5.4-mini}): is it a single question (with an optional informational note on the form progress)?

\textbf{No Acknowledgement} (\texttt{openai/gpt-5.5}): does it avoid generating any form of acknowledgement for the previous user answer?

\textbf{No Value Echo} (\texttt{openai/gpt-5.5}): does it avoid reading the captured value back to the caller?
\end{quote}

The full prompt for each is given in the example below. Only the ``Correctness'' prompt varies by test case; the other four are the same for every reply.

\paragraph{Example.} A \replyllm{} test case is shown below, with the same conventions.

\begin{quote}
\textbf{Conversation history}

\textbf{assistant:} Namaste! Yah ek automated call hai, ek nishulk maatru evam shishu swasthya seva ki taraf se. Aapne hamari seva se judne ke liye Anganwadi se sampark kiya hai. Yah ek muft seva hai, jismein aapko maa aur bachche ke swasthya se judi upyogi jaankari milegi. Hamari seva se judne ke liye hum aapse kuch sawal poochhenge. Kripya unke jawab bolkar dein. Aapki di gayi saari jaankari hamare saath surakshit rahegi. Apna poora naam bataiye \emph{(Namaste! This is an automated call from a free maternal
and child health service. You contacted the Anganwadi to join our service. This is a free service through which you will receive useful information about mother and child health. To enroll you in our service we will ask you a few questions. Please answer them by speaking. All the information you give will stay safe with us. Please tell me your full name.)}

\textbf{user:} Mera poora naam Mrinmayee Kshirsagar hai \emph{(My full name is Mrinmayee Kshirsagar.)}

\textbf{assistant:} Theek hai. Apne zile ka naam bataayein \emph{(All right. Please tell me the name of your district.)}

\textbf{user:} Form ke liye, mera zila Kamrup Metropolitan hai. \emph{(For the form, my district is Kamrup Metropolitan.)}

\textbf{assistant:} Theek hai. Apne Anganwadi ka naam bataayein \emph{(All right. Please tell me the name of your Anganwadi.)}

\textbf{user:} Mere aanganwadi ka naam Ghorakshpalli hai \emph{(The name of my Anganwadi is Ghorakshpalli.)}

\textbf{assistant:} Theek hai. Kya aap abhi garbhvati hain? \emph{(All right. Are you currently pregnant?)}

\textbf{user:} Haan main abhi garbhvati hoon. \emph{(Yes, I am currently pregnant.)}

\textbf{tool:} \texttt{validate\_user\_turn}

\textbf{tool output:} Correct value \emph{True} captured for the pregnancy field. Pending questions: (1)~gestational age in months, optional, integer, script ``Aapko garbhvati hue kitne mahine hue hain?'' \emph{(How many months pregnant are you?)}; the answer is a number of months.

\medskip
\textbf{Evaluation.} The reply the agent produces next is sent to all five judges. Each judge receives the conversation above and their respective system prompts.

\medskip
\emph{Correctness.} You are a highly accurate evaluator evaluating the response of an agent to a user's message. You will be given a conversation between a user and an agent along with the response of the agent to the final user message. You need to evaluate if the response adheres to the evaluation criteria: The reply advances the interview to the correct next question: the one whose script is ``Aapko garbhvati hue kitne mahine hue hain?'' \emph{(How many months pregnant are you?)}. It asks that specific question (not an earlier, later, or invented one), does not re-ask an already-answered question, and does not end the call.

\medskip
\emph{Hindi Language adherence.} You are a highly accurate evaluator checking the language of an agent's reply. You will be given a conversation between a user and an agent along with the agent's response to the final user message. You need to evaluate if the last agent response adheres to the evaluation criteria: The last agent reply should be largely written in Hindi (Devanagari script). It does not switch to roman alphabets or english words except for using proper nouns and everyday common-use english words in devnagari (e.g. saying ``Whatsapp'' as its Devanagari transliteration is fine) or digits in roman numerals are fine. If you think there has been any violation, give concrete examples of the violation in your reasoning.

\medskip
\emph{Reply Conciseness.} You are a highly accurate evaluator checking whether an agent's reply is concise. You will be given a conversation between a user and an agent along with the agent's response to the final user message. You need to evaluate if the response adheres to the evaluation criteria: The reply is concise: it asks exactly ONE question, optionally preceded by the ONE informational preamble the script requires for this field. It does not bundle multiple questions, enumerate pending fields, summarise progress, or add chit-chat beyond what the script requires.

\medskip
\emph{No Acknowledgement.} You are a highly accurate evaluator checking whether an agent's reply is free of any acknowledgement of the user's previous answer. You will be given a conversation between a user and an agent along with the agent's response to the final user message. You need to evaluate if the response adheres to the evaluation criteria: The reply begins directly with its substantive line. Examples: the question, the re-ask, or a required informational/skip script like stating the progress update of the form completion so far, etc. It contains NO acknowledgement of the user's previous answer like one or more of the following but not strictly limited to them: no receipt, back-channel, reassurance, praise, or transition phrase in any language (e.g.\ ``it's okay'', ``okay'', ``yes'', ``alright'', ``got it'', ``thank you'', ``no problem'').

\medskip
\emph{No Value Echo.} You are a highly accurate evaluator checking whether an agent's reply avoids echoing the user's just-provided value. You will be given a conversation between a user and an agent along with the agent's response to the final user message. You need to evaluate if the response adheres to the evaluation criteria: The reply does not read back, repeat, or confirm the specific value the user just provided for the field that was just captured (e.g. echoing their name/number/DOB etc. back at them as confirmation). Asking the next scripted question, or speaking a required verbatim re-ask line, even one containing a quoted example token, is NOT an echo and must pass. The chat history will contain a tool call prior to the last agent message. You are not supposed to check the agent message against that tool call message. You need to compare the last agent message with only the user messages prior to it.
\end{quote}

\subsection{Calibration Procedure}
\label{app:judge-calibration}

Every judge was calibrated against human labels before being used to score the benchmark. We drew a batch of 50 unit tests, ran each judge over them, and had the authors independently label every judgment. The initial prompts were not fully aligned: each judge disagreed with the human label on some cases. We revised the prompts over several iterations until every judge agreed with the human labels on all 50 tests. We then applied the final prompts unchanged to a held-out batch of 50 tests that had played no part in the iteration, on which the LLM judge outputs matched the human labels on every case. We acknowledge that the size of the calibration dataset is small and plan to expand it in future work.

\section{Full Results with Confidence Intervals}
\label{app:results}

Tables~\ref{app:ci-stt}--\ref{app:ci-response} show the model comparison results for speech-to-text (STT), extraction accuracy, form completion and response accuracy with 95\% confidence intervals.

\begin{table}[!ht]
\centering
\small
\setlength{\tabcolsep}{3pt}
\begin{tabular}{lrrr}
\hline
Model & WER $\downarrow$ & LLM-WER $\downarrow$ & Cost (\$) $\downarrow$ \\
\hline
Scribe v2 & $0.835 \pm 0.015$ & $0.062 \pm 0.024$ & $0.172$ \\
Chirp 3 & $0.981 \pm 0.007$ & $\mathbf{0.055 \pm 0.025}$ & $0.424$ \\
Saaras v3 & $1.007 \pm 0.004$ & $0.074 \pm 0.028$ & $0.137$ \\
GPT-4o-transcr. & $1.012 \pm 0.004$ & $0.127 \pm 0.033$ & $0.159$ \\
Nova-3 & $\mathbf{0.826 \pm 0.015}$ & $0.104 \pm 0.031$ & $\mathbf{0.127}$ \\
\hline
\end{tabular}
\caption{Comparison of STT models. Since models have different billing units, the total cost (USD) of transcribing the entire benchmark is reported.}
\label{app:ci-stt}
\end{table}

\begin{table*}[!t]
\centering
\small
\setlength{\tabcolsep}{3pt}
\begin{tabular}{l cccc}
\hline
Model & Reference & Saaras v3 & Scribe v2 & Nova-3\\
\hline
GPT-5.5 & $\mathbf{99.79 \pm 0.99}$ & $95.53 \pm 1.03$ & $98.62 \pm 0.64$ & $95.16 \pm 1.07$\\
Gemini 3.5 Flash & $99.36 \pm 1.22$ & $95.69 \pm 1.01$ & $\mathbf{98.94 \pm 0.58}$ & $\mathbf{96.38 \pm 0.94}$\\
Claude Opus 4.8 & $99.15 \pm 1.32$ & $\mathbf{96.22 \pm 0.96}$ & $98.40 \pm 0.67$ & $96.01 \pm 0.98$\\
Claude Sonnet 4.6 & $99.15 \pm 1.32$ & $96.01 \pm 0.98$ & $\mathbf{98.94 \pm 0.58}$ & $95.96 \pm 0.99$\\
GLM-5.1 & $98.72 \pm 1.48$ & $94.52 \pm 1.12$ & $63.40 \pm 2.20$ & $95.43 \pm 1.05$\\
Gemini Pro & $97.23 \pm 1.90$ & $94.26 \pm 1.15$ & $97.29 \pm 0.84$ & $94.63 \pm 1.12$\\
Gemini 2.5 Flash & $96.17 \pm 2.14$ & $95.27 \pm 1.06$ & $98.03 \pm 0.73$ & $95.43 \pm 1.05$\\
Gemini 3 Flash & $95.96 \pm 2.19$ & $84.79 \pm 1.70$ & $88.83 \pm 1.50$ & $86.28 \pm 1.63$\\
GPT-5.4-mini & $92.34 \pm 2.76$ & $90.05 \pm 1.43$ & $96.76 \pm 0.91$ & $91.54 \pm 1.34$\\
Mistral Medium 3.5 & $89.36 \pm 3.11$ & $87.82 \pm 1.56$ & $92.34 \pm 1.29$ & $87.77 \pm 1.56$\\
GPT-4.1 & $85.11 \pm 3.51$ & $75.53 \pm 1.99$ & $78.94 \pm 1.91$ & $79.84 \pm 1.87$\\
\hline
\end{tabular}
\caption{Per-turn extraction accuracy using reference transcripts and the outputs of the three STT models as inputs.}
\label{app:ci-extraction}
\end{table*}

\begin{table*}[!t]
\centering
\small
\setlength{\tabcolsep}{3pt}
\begin{tabular}{l cccc}
\hline
Model & Reference & Saaras v3 & Scribe v2 & Nova-3\\
\hline
Gemini 3 Flash & $\mathbf{100.00 \pm 0.00}$ & $85.24 \pm 0.77$ & $89.48 \pm 0.60$ & $83.92 \pm 1.01$\\
Gemini 3.5 Flash & $\mathbf{100.00 \pm 0.00}$ & $90.74 \pm 0.61$ & $92.50 \pm 0.53$ & $\mathbf{87.09 \pm 0.76}$\\
GPT-5.5 & $99.95 \pm 0.09$ & $89.97 \pm 0.73$ & $89.37 \pm 0.64$ & $84.73 \pm 0.80$\\
Claude Sonnet 4.6 & $99.66 \pm 0.23$ & $91.55 \pm 0.71$ & $93.01 \pm 0.50$ & $84.54 \pm 0.77$\\
GLM-5.1 & $99.56 \pm 0.27$ & $90.92 \pm 0.75$ & $58.66 \pm 1.46$ & $86.09 \pm 0.75$\\
Claude Opus 4.8 & $99.41 \pm 0.31$ & $92.22 \pm 0.61$ & $92.03 \pm 0.56$ & $85.83 \pm 0.88$\\
Gemini 2.5 Flash & $99.06 \pm 0.38$ & $91.76 \pm 0.70$ & $92.39 \pm 0.52$ & $85.50 \pm 0.73$\\
Gemini Pro & $98.90 \pm 0.42$ & $\mathbf{92.55 \pm 0.52}$ & $\mathbf{93.47 \pm 0.45}$ & $86.53 \pm 0.70$\\
GPT-5.4-mini & $98.35 \pm 0.48$ & $90.68 \pm 0.75$ & $90.10 \pm 0.55$ & $84.40 \pm 0.79$\\
GPT-4.1 & $95.99 \pm 0.70$ & $88.82 \pm 0.74$ & $91.50 \pm 0.53$ & $86.33 \pm 0.74$\\
Mistral Medium 3.5 & $92.83 \pm 0.68$ & $90.39 \pm 0.68$ & $91.84 \pm 0.49$ & $83.37 \pm 0.76$\\
\hline
\end{tabular}
\caption{End-to-end form-completion accuracy using reference transcripts and the outputs of the three STT models as inputs.}
\label{app:ci-form-completion}
\end{table*}

\begin{table*}[!t]
\centering
\small
\setlength{\tabcolsep}{4pt}
\begin{tabular}{@{}lcccc@{}}
\toprule
Model & \multicolumn{2}{c}{Response accuracy} & Latency (ms) & Cost (\$/turn) \\
\cmidrule(lr){2-3}
& Reference & Scribe v2 & & \\
\midrule
Claude Sonnet 4.6 & $\mathbf{100.00 \pm 0.00}$ & $\mathbf{97.77 \pm 0.75}$ & $5148 \pm 392$ & $0.0146 \pm 0.0000$ \\
GPT-4.1 & $\mathbf{100.00 \pm 0.00}$ & $96.60 \pm 0.91$ & $3647 \pm 656$ & $0.0025 \pm 0.0001$ \\
GPT-5.4-mini & $98.09 \pm 1.28$ & $96.70 \pm 0.85$ & $3356 \pm 523$ & $\mathbf{0.0008 \pm 0.0000}$ \\
Gemini 3 Flash & $97.23 \pm 1.49$ & $96.17 \pm 0.96$ & $\mathbf{2661 \pm 239}$ & $0.0017 \pm 0.0000$ \\
Gemini 3.5 Flash & $97.02 \pm 1.28$ & $94.73 \pm 1.11$ & $3496 \pm 1322$ & $0.0052 \pm 0.0001$ \\
\bottomrule
\end{tabular}
\caption{\replyllm{} response accuracy using reference transcripts and transcripts from Scribe v2 as inputs.}
\label{app:ci-response}
\end{table*}

\section{Model Selection}
\label{app:selection}

\subsection{\extractllm{} Selection}
\label{app:selection-extract}

Table~\ref{app:scalarization-extract} gives the optimal \extractllm{} model using the weighted-sum scalarization method for $0.50 \leq w_a \leq 0.90$ with $w_\ell=0.1$.

\begin{table*}[!t]
\centering
\small
\setlength{\tabcolsep}{3pt}
\begin{tabular}{rrlr@{\hspace{0.6em}}rrlr@{\hspace{0.6em}}rrlr}
\hline
$w_a$ & $w_c$ & Model & $U$ & $w_a$ & $w_c$ & Model & $U$ & $w_a$ & $w_c$ & Model & $U$ \\
\hline
0.50 & 0.40 & Gemini 3.5 Flash & 0.763 & 0.64 & 0.26 & Sonnet 4.6 & 0.776 & 0.78 & 0.12 & Sonnet 4.6 & 0.843 \\
0.51 & 0.39 & Gemini 3.5 Flash & 0.763 & 0.65 & 0.25 & Sonnet 4.6 & 0.781 & 0.79 & 0.11 & Sonnet 4.6 & 0.848 \\
0.52 & 0.38 & Gemini 3.5 Flash & 0.763 & 0.66 & 0.24 & Sonnet 4.6 & 0.786 & 0.80 & 0.10 & Sonnet 4.6 & 0.852 \\
0.53 & 0.37 & Gemini 3.5 Flash & 0.763 & 0.67 & 0.23 & Sonnet 4.6 & 0.791 & 0.81 & 0.09 & Sonnet 4.6 & 0.857 \\
0.54 & 0.36 & Gemini 3.5 Flash & 0.763 & 0.68 & 0.22 & Sonnet 4.6 & 0.795 & 0.82 & 0.08 & Sonnet 4.6 & 0.862 \\
0.55 & 0.35 & Gemini 3.5 Flash & 0.763 & 0.69 & 0.21 & Sonnet 4.6 & 0.800 & 0.83 & 0.07 & Sonnet 4.6 & 0.867 \\
0.56 & 0.34 & Gemini 3.5 Flash & 0.764 & 0.70 & 0.20 & Sonnet 4.6 & 0.805 & 0.84 & 0.06 & Sonnet 4.6 & 0.871 \\
0.57 & 0.33 & Gemini 3.5 Flash & 0.764 & 0.71 & 0.19 & Sonnet 4.6 & 0.810 & 0.85 & 0.05 & Sonnet 4.6 & 0.876 \\
0.58 & 0.32 & Gemini 3.5 Flash & 0.764 & 0.72 & 0.18 & Sonnet 4.6 & 0.814 & 0.86 & 0.04 & Sonnet 4.6 & 0.881 \\
0.59 & 0.31 & Gemini 3.5 Flash & 0.764 & 0.73 & 0.17 & Sonnet 4.6 & 0.819 & 0.87 & 0.03 & Sonnet 4.6 & 0.886 \\
0.60 & 0.30 & Gemini 3.5 Flash & 0.764 & 0.74 & 0.16 & Sonnet 4.6 & 0.824 & 0.88 & 0.02 & Sonnet 4.6 & 0.890 \\
0.61 & 0.29 & Gemini 3.5 Flash & 0.764 & 0.75 & 0.15 & Sonnet 4.6 & 0.829 & 0.89 & 0.01 & Sonnet 4.6 & 0.895 \\
0.62 & 0.28 & Sonnet 4.6 & 0.767 & 0.76 & 0.14 & Sonnet 4.6 & 0.833 & 0.90 & 0.00 & Sonnet 4.6 & 0.900 \\
0.63 & 0.27 & Sonnet 4.6 & 0.772 & 0.77 & 0.13 & Sonnet 4.6 & 0.838 &  & & &  \\
\hline
\end{tabular}
\caption{Optimal \extractllm{} model under weighted-sum scalarization for
$0.50 \leq w_a \leq 0.90$, with $w_\ell=0.1$ and $w_c=1-w_a-w_\ell$, and its composite score $U$.}
\label{app:scalarization-extract}
\end{table*}

\subsection{\replyllm{} Selection}
\label{app:selection-reply}

\begin{figure}[!ht]
  \centering
  \includegraphics[width=\columnwidth]{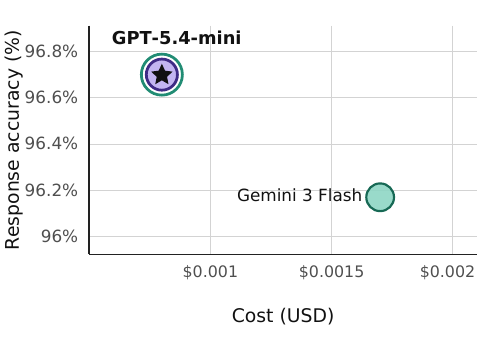}
\caption{Cost--quality--latency trade-off among \replyllm{} models satisfying the deployment constraints. The logarithmic x-axis shows cost per turn, the y-axis shows response accuracy using Scribe v2 transcripts, and marker area encodes p95 latency per turn (bigger is slower). Both models are Pareto-optimal across the three objectives. The model selected for deployment is highlighted.}
  \label{fig:reply-pareto}
\end{figure}

GPT-5.4-mini is optimal throughout $0.50 \leq w_a \leq 0.90$ ($w_\ell=0.1$), among the models on the Pareto frontier (Figure~\ref{fig:reply-pareto}), with a consistent composite score of $U=0.9$ across the sweep. 

\section{Implementation Details}
\label{app:implementation}

\paragraph{Orchestration and telephony.} We use Pipecat \citep{pipecat} for
orchestration and Exotel \citep{exotel} for telephony, which delivers 8\,kHz $\mu$-law audio
over a WebSocket. For VAD, we use Silero \citep{silerovad} at the same 8\,kHz sample rate.
It marks the start of speech after $0.1$\,s, the end after $0.2$\,s of
silence, and ignores any audio that scores below $0.7$ confidence or $0.6$ loudness. The agent waits a further 0.4 s after that before marking the turn as completed, so a caller pausing mid-answer is not cut off. If the caller has not spoken for more than 3.0 s since the agent stopped speaking, we re-prompt. Each audio file in the benchmark is 16\,kHz mono 16-bit PCM.

\paragraph{Models.} All LLM calls are served through OpenRouter \citep{openrouter}. We
set the temperature to $0$ for non-reasoning models, and the reasoning effort to ``medium'' for \extractllm{} and ``low'' for \replyllm{}. For both, we cap the
output at $16{,}000$ tokens and pass the most recent 200 turns of the
conversation as input. For TTS, we use Google Cloud Chirp 3 HD \citep{chirp3hd} with the female Hindi voice
\texttt{Achernar}, slowed to $0.9\times$ the default speed so the questions
are easier to follow.

\paragraph{Evaluation harness.} All STT and LLM evaluations were run using Calibrate \citep{calibrate}.

\paragraph{Environment.} Experiments were run from a MacBook Pro (Apple M4
Pro, $24$\,GB) on macOS 15.7 with Python 3.11, using \texttt{pipecat-ai}
1.2.1, \texttt{openai} 2.15.0, \texttt{instructor} 1.13.0, \texttt{pydantic}
2.12.3, \texttt{jiwer} \citep{jiwer} 4.0.0, \texttt{indic-nlp-library} \citep{indicnlp} 0.92,
\texttt{pydub} 0.25.1 and \texttt{numpy} 2.2.6.

\paragraph{Determinism.} No random seeds were set: every model is served by a
hosted API that offers no determinism guarantee, so identical settings can
still produce different outputs. Each configuration was evaluated
once over the full test suite, so the confidence intervals in
Section~\ref{app:results} reflect variation across test items rather than
across repeated runs.

\end{document}